\documentclass[conference]{IEEEtran}
\IEEEoverridecommandlockouts

\usepackage{cite}
\usepackage{amsmath,amssymb,amsfonts}
\usepackage{algorithmic}
\usepackage{graphicx}
\usepackage{textcomp}
\usepackage{xcolor}
\usepackage{url}
\usepackage{hyperref}
\hypersetup{hidelinks}
\usepackage{tikz}
\usetikzlibrary{positioning, arrows.meta, calc}
\def\BibTeX{{\rm B\kern-.05em{\sc i\kern-.025em b}\kern-.08em
    T\kern-.1667em\lower.7ex\hbox{E}\kern-.125emX}}
\begin{document}

\title{\fontsize{21}{24}\selectfont Structured Clinical Memory as Context for Probabilistic Decision Models: A Synthetic Evaluation with Jev}

\author{\IEEEauthorblockN{Jash Sayani}
\IEEEauthorblockA{\href{mailto:jash.sayani@abbycare.ai}{\texttt{jash.sayani@abbycare.ai}}}
\and
\IEEEauthorblockN{Gaurang Karwande}
\IEEEauthorblockA{\href{mailto:gaurang.karwande@abbycare.ai}{\texttt{gaurang.karwande@abbycare.ai}}}
\and
\IEEEauthorblockN{Achinth Bharadwaj}
\IEEEauthorblockA{\href{mailto:achinth.bharadwaj@abbycare.ai}{\texttt{achinth.bharadwaj@abbycare.ai}}}
}
\maketitle

\begin{abstract}
Longitudinal clinical context is often distributed across noisy, repetitive forms and narrative records. We study whether a compact structured memory can improve a probabilistic decision model's answers when the underlying patient vignette is already available. We created 30 fully synthetic cases spanning twelve clinical domains, paired each with synthetically generated structured memory, and evaluated 120 typed questions with Jev. The raw-vignette arm achieved 90.0\% accuracy (95\% normal-approximation confidence interval: 84.6\%-95.4\%); adding memory achieved 93.3\% (88.9\%-97.8\%). Five questions changed from incorrect to correct and one changed in the opposite direction (exact McNemar $p=0.219$). Choice log loss decreased from 0.258 to 0.175 and binary Brier score from 0.054 to 0.022. Memory added 310 input tokens per call on average, while observed mean latency was unchanged at approximately 0.5 seconds. These results are directionally positive rather than clinically validating: the cases are synthetic, the labels are single-author judgments, and the memory is an oracle upper bound rather than extracted production output. The pilot supports evaluating structured clinical memory as a reusable state interface for decision models.
\end{abstract}


\section{Structured context may improve decision models}

Clinical decision support frequently depends on facts recorded outside the immediate encounter: an active condition, a resolved episode, an uncertain diagnosis, a medication, or a prior contraindication. Repeatedly passing all underlying forms to a model preserves source detail but also repeats irrelevant text and asks the model to reconstruct longitudinal state on every call.

We study a narrower interface: a maintained clinical memory containing typed features such as condition or medication, status, onset, certainty, source form, negation, and a short note. This memory is not a free-form summary. It is structured state that can be appended to a current vignette and consumed by a model whose outputs are choices, probability weighted choices, binary judgments, or ordinal scores.

The concrete model in this pilot is Jev, described by its developer TypeSafe AI as accepting unstructured state and returning typed probabilistic decisions in parallel [1]. The study does not test the vendor's broader performance claims. It asks whether the quality and accuracy of the responses change when structured clinical memory accompanies the raw vignette.

\subsection*{Contributions}

This pilot contributes: (i) a paired benchmark of 30 synthetic cases and 120 typed questions; (ii) an explicit memory representation capturing longitudinal status and provenance; (iii) a two-arm evaluation using identical questions; and (iv) accuracy, probability-quality, confidence, token, and latency analyses with limitations stated as part of the result.

\section{Related work}

External-memory systems for language models manage information beyond a single context window. MemGPT, for example, frames memory as a hierarchy whose contents can be moved between limited context and external storage [2]. Our study differs in both representation and consumer: the memory is a fixed clinical feature schema, and the downstream model emits typed probabilistic decisions rather than generated dialogue.

Clinical natural-language processing has long extracted concepts and attributes from records. cTAKES introduced a modular pipeline for clinical free-text annotations, including concept mapping, negation, and status attributes [3]. This pilot does not evaluate extraction. Its synthetically generated memory intentionally isolates the downstream question: if a suitable structured state exists, does it change decision quality? We intentionally isolate agent memory as an independent component in order to isolate and understand its quality contribution to decision quality given a structured state.

Probability quality matters because workflows do not just consume the decision, but also the uncertainty of the decisions. We therefore report log loss for multiple-choice outputs and Brier score for binary probabilities [4], alongside thresholded accuracy.

\section{The benchmark isolates context representation}

\subsection{Synthetic cases and questions}

We authored 30 synthetic patient cases across twelve domain labels: cardiology, diabetes, respiratory, infectious disease, neurology, mental health, musculoskeletal, renal, gastrointestinal, allergy, geriatrics, and pain. No real patient records or protected health information were used. Each vignette intentionally included clinically relevant facts plus ordinary ambiguity or distractors.

The benchmark contains 120 questions: 56 multiple-choice questions, 36 binary questions, and 28 five-level ordinal scores. Tasks cover triage priority, next action, risk scoring, condition presence, and contraindication checks. Gold answers and rationales were authored once and shared by both experimental arms.

\subsection{Structured clinical memory}

Each case has a synthetically generated memory comprising a list of condition or medication features. Every feature includes a normalized label, longitudinal status (\textit{active}, \textit{history\_of}, or \textit{resolved}), onset, certainty, source form, negation flag, and optional note. The memory was designed to preserve facts that are often obscured when several forms are read together.

\subsection{Paired experimental arms}

Every case was evaluated twice. Arm A supplied the synthetic vignette as model state. Arm B supplied the same vignette followed by the structured clinical-memory block. All questions for one case were submitted in a single call and evaluated in parallel. Thus, 30 cases multiplied by two arms yielded 60 model calls.

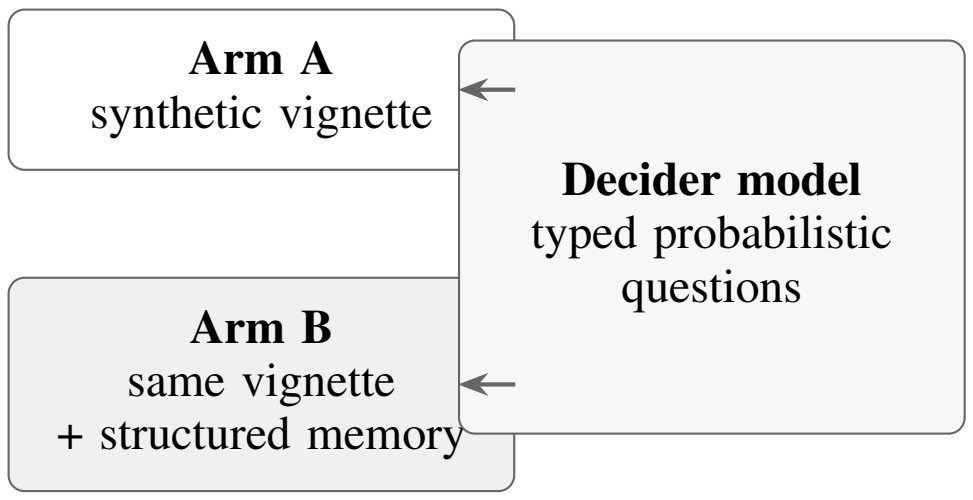
\begin{figure}[t]
\centering
\begin{tikzpicture}[
  box/.style={draw=black!60, rounded corners=3pt, align=center,
              inner sep=6pt, text width=0.33\columnwidth, font=\small},
  armB/.style={box, fill=black!6},
  decider/.style={box, fill=black!3, minimum height=26mm},
  arr/.style={-{Stealth[length=2mm]}, thick, draw=black!60}
]
\node[box] (A) {\textbf{Arm A}\\ synthetic vignette};
\node[armB, below=7mm of A] (B) {\textbf{Arm B}\\ same vignette\\ + structured memory};
\node[decider] (D) at ($(A.east)!0.5!(B.east) + (13mm, 0)$)
  {\textbf{Decider model}\\ typed probabilistic\\ questions};
\draw[arr] (A.east) -- (D.west |- A.east);
\draw[arr] (B.east) -- (D.west |- B.east);
\end{tikzpicture}
\caption{Paired evaluation design. Only the presence of structured
clinical memory differs between the two arms; the vignette, decider
model, and typed probabilistic questions are identical.}
\label{fig:design}
\end{figure}

\section{Outcomes and analysis}

\subsection{Primary correctness metrics}

Multiple-choice answers were correct only when the selected option matched the gold key. Binary outputs were mapped to ``yes'' when the reported probability was at least 0.5. Ordinal score outputs were correct when their probability-weighted value was within one level of the gold score.

\begin{table}[t]
\centering
\small
\setlength{\tabcolsep}{9pt}
\renewcommand{\arraystretch}{1.4}
\begin{tabular}{|l|l|l|}
\hline
\textbf{Output} & \textbf{Accuracy rule} & \textbf{Probability metric} \\
\hline
Choice & Exact option & Log loss \\
\hline
Binary & P(yes) $\geq$ 0.5 & Brier score \\
\hline
Score & Absolute error $\leq$ 1 & Mean absolute error \\
\hline
\end{tabular}
\end{table}

\subsection{Paired inference}

Overall accuracy uses all 120 questions in each arm. We report normal-approximation 95\% confidence intervals for descriptive continuity with the scoring harness. Because each question appears in both arms, we additionally report the discordant-pair counts and a two-sided exact McNemar test. With only six discordant pairs, the test has low power.

\subsection{Confidence and efficiency}

For confidence analysis, predictions were grouped into five bins from 0.5 to 1.0. We compare mean confidence and empirical accuracy within each populated bin, while noting that some outputs lacked a confidence field and were excluded from this analysis. Efficiency outcomes are mean input tokens and wall-clock latency per call as reported by the run harness.

\subsection{Scope}

The unit of analysis is a model answer, not a patient outcome. No clinicians reviewed the labels, no safety endpoint was measured, and no intervention was recommended or delivered. The experiment is a controlled software benchmark.

\section{Accuracy improved by 3.3 percentage points}

Arm A answered 108 of 120 questions correctly (90.0\%). Arm B answered 112 correctly (93.3\%), an absolute difference of 3.3 percentage points. The descriptive confidence intervals overlap. In the paired analysis, five questions changed from incorrect in Arm A to correct in Arm B, while one changed from correct to incorrect. The resulting exact two-sided McNemar p-value is 0.219.

\begin{table}[t]
\centering
\small
\setlength{\tabcolsep}{9pt}
\renewcommand{\arraystretch}{1.4}
\begin{tabular}{|l|c|c|c|}
\hline
\textbf{Condition} & \textbf{Correct} & \textbf{Accuracy} & \textbf{95\% CI} \\
\hline
Arm A: vignette & 108 / 120 & 90.0\% & 84.6\%--95.4\% \\
\hline
Arm B: + memory & 112 / 120 & 93.3\% & 88.9\%--97.8\% \\
\hline
Difference & +4 & +3.3 pp & Not estimated \\
\hline
\end{tabular}
\end{table}

This pattern is compatible with a modest benefit, but it is not sufficient to reject the null of equal paired error rates. The small number of errors and discordant pairs dominates the uncertainty.

\subsection{Gains appeared across output types}

\begin{table}[t]
\centering
\small
\setlength{\tabcolsep}{9pt}
\renewcommand{\arraystretch}{1.4}
\begin{tabular}{|l|c|c|c|}
\hline
\textbf{Question type} & \textbf{n} & \textbf{Arm A} & \textbf{Arm B} \\
\hline
Choice accuracy & 56 & 89.3\% & 91.1\% \\
\hline
Binary accuracy & 36 & 94.4\% & 97.2\% \\
\hline
Score within $\pm$1 & 28 & 85.7\% & 92.9\% \\
\hline
\end{tabular}
\end{table}

The largest descriptive increase was on ordinal scores: 7.2 percentage points. However, the per-type samples are small and their confidence intervals overlap. These breakdowns should be treated as hypotheses for a larger benchmark rather than stable estimates of task-specific effect.

\section{Probability quality improved on choice and binary tasks}

Choice log loss decreased from 0.258 to 0.175, indicating that the memory condition assigned more probability to the gold option on average. Binary Brier score decreased from 0.054 to 0.022. For ordinal scores, mean absolute error decreased from 0.521 to 0.469 levels.

\begin{table}[t]
\centering
\small
\setlength{\tabcolsep}{9pt}
\renewcommand{\arraystretch}{1.4}
\begin{tabular}{|l|c|c|c|}
\hline
\textbf{Metric (lower is better)} & \textbf{Arm A} & \textbf{Arm B} & \textbf{Change} \\
\hline
Choice log loss & 0.258 & 0.175 & $-$0.083 \\
\hline
Binary Brier score & 0.054 & 0.022 & $-$0.032 \\
\hline
Score MAE & 0.521 & 0.469 & $-$0.052 \\
\hline
\end{tabular}
\end{table}

\subsection{Memory increased high-confidence predictions while preserving their accuracy}

The 0.9-1.0 confidence bin contained 55 Arm A predictions with 100\% empirical accuracy and 65 Arm B predictions with 98.5\% accuracy. Memory therefore moved more predictions into the highest-confidence bin while preserving high observed accuracy. Other bins were sparse and non-monotonic, so the pilot does not establish formal calibration.

\begin{table}[t]
\centering
\small
\setlength{\tabcolsep}{9pt}
\renewcommand{\arraystretch}{1.4}
\begin{tabular}{|l|c|c|c|c|}
\hline
\textbf{Confidence bin} & \multicolumn{2}{c|}{\textbf{Arm A}} & \multicolumn{2}{c|}{\textbf{Arm B}} \\
\hline
 & \textbf{n} & \textbf{acc.} & \textbf{n} & \textbf{acc.} \\
\hline
0.5--0.6 & 9 & 88.9\% & 8 & 87.5\% \\
\hline
0.6--0.7 & 12 & 91.7\% & 8 & 100\% \\
\hline
0.7--0.8 & 23 & 78.3\% & 18 & 88.9\% \\
\hline
0.8--0.9 & 16 & 81.2\% & 14 & 100\% \\
\hline
0.9--1.0 & 55 & 100\% & 65 & 98.5\% \\
\hline
\end{tabular}
\end{table}

\section{Memory added tokens but not observed latency}

Mean input length increased from 553.9 to 863.9 tokens per call, an increment of 310.0 tokens (56.0\%). Mean measured latency was 0.51 seconds in Arm A and 0.50 seconds in Arm B across 30 calls per arm.

\begin{table}[t]
\centering
\small
\setlength{\tabcolsep}{9pt}
\renewcommand{\arraystretch}{1.4}
\begin{tabular}{|l|c|c|}
\hline
\textbf{Efficiency outcome} & \textbf{Arm A} & \textbf{Arm B} \\
\hline
Mean input tokens & 553.9 & 863.9 \\
\hline
Mean latency & 0.51 s & 0.50 s \\
\hline
Calls & 30 & 30 \\
\hline
\end{tabular}
\end{table}

The near-equal latency should not be interpreted as a general performance guarantee. The run was small, networked, and not designed for latency inference. The token result is more direct: this implementation trades additional structured context for a modest accuracy and probability-quality gain.

\section{The result is an upper-bound pilot}

The main finding is not that structured memory solves clinical reasoning. It is that, under a controlled synthetic benchmark and a strong raw-text baseline, appending a compact longitudinal state was associated with four additional correct answers and better probability scores. The direction is consistent across choice, binary, and ordinal outputs, but statistical uncertainty remains substantial.

\subsection{Why oracle memory matters}

Arm B receives memory that is cleaner than a production extraction pipeline would provide. Real systems may omit a condition, merge two entities incorrectly, mishandle negation, preserve stale status, or assign unjustified certainty. Any of these errors could reduce or reverse the observed benefit. The result should therefore be read as a ceiling for this representation on this benchmark, not an estimate of deployed performance.

\subsection{The feature schema is the intervention}

The memory block does more than shorten text. It asserts a longitudinal interpretation: which features are active, historical, resolved, negated, or uncertain, and which source form supports each item. The evaluation cannot separate the contribution of structure from the contribution of expert-authored normalization. Future ablations should compare the full schema with simpler summaries and retrieval of raw source passages.

\section{Limitations}

\begin{enumerate}
\item \textbf{Synthetic data.} The cases do not reproduce the distribution, ambiguity, documentation errors, or missingness of real clinical records.
\item \textbf{Single-author gold labels.} Labels and rationales were not independently reviewed or adjudicated by clinicians.
\item \textbf{Oracle memory.} The study does not measure extraction precision, recall, deduplication, temporal updating, or failure propagation.
\item \textbf{Small sample.} With 120 questions and six discordant pairs, the paired comparison is underpowered.
\item \textbf{Question dependence.} Four questions share each case, so treating all 120 as independent understates clustering in descriptive intervals.
\item \textbf{One model and one run.} Results may vary across model revisions, sampling behavior, service conditions, and alternative decision architectures.
\item \textbf{Thresholded metrics.} Binary accuracy depends on a 0.5 threshold, while ordinal correctness allows $\pm$1 level; alternate decision costs may favor different rules.
\item \textbf{No clinical endpoint.} Benchmark correctness does not establish safety, utility, equity, or improved patient outcomes.
\end{enumerate}

\section{Next evaluation must test extracted memory prospectively}

A stronger study should connect representation quality to downstream decisions. First, a frozen extraction system should produce memory from a preregistered set of de-identified or fully synthetic forms. Clinicians should independently label features and adjudicate disagreements, enabling precision, recall, F1, and temporal-status error analysis.

Second, the paired decision benchmark should be expanded and grouped by patient, with sample size chosen for a clinically meaningful effect. Analyses should use cluster-aware intervals or patient-level bootstrap resampling. Errors should be stratified by task, specialty, memory dependence, and potential harm.

Third, comparisons should include raw forms, retrieved source passages, unstructured summaries, and structured memory. Schema ablations can test whether status, onset, certainty, source provenance, or negation produces the gain. A deliberately corrupted-memory arm would quantify sensitivity to stale or incorrect features.

Finally, a prospective human-factors evaluation should measure whether clinicians find the resulting decisions appropriately calibrated and whether the memory makes source verification easier. Human oversight, auditability, and abstention behavior are central for any clinical use.

\section{Ethical and safety considerations}

This study uses synthetic cases only. The reported probabilities and decisions are benchmark outputs, not medical advice. The benchmark contains no demographic fairness analysis and cannot support claims about subgroup performance. Before use with real records, data governance, privacy controls, access logging, source traceability, bias evaluation, and clinical review would be required.

\section{Conclusion}

Structured clinical memory is associated with increased accuracy from 90.0\% to 93.3\% and improved probability scores in a small synthetic Jev benchmark. The result is directionally encouraging but not statistically conclusive or clinically validating. Its value is as a testable research hypothesis: a maintained, typed account of longitudinal clinical state may be a useful context interface for probabilistic decision models. The next step is to replace oracle memory with measured extraction output and evaluate the full error chain under clinician review.

\appendix

\section{Benchmark composition}

\begin{table}[h]
\centering
\small
\setlength{\tabcolsep}{9pt}
\renewcommand{\arraystretch}{1.4}
\begin{tabular}{|l|c|c|}
\hline
\textbf{Domain label} & \textbf{Cases} & \textbf{Questions} \\
\hline
Cardiology & 3 & 12 \\
\hline
Diabetes & 3 & 12 \\
\hline
Respiratory & 3 & 12 \\
\hline
Infectious disease & 3 & 12 \\
\hline
Neurology & 3 & 12 \\
\hline
Mental health & 3 & 12 \\
\hline
Musculoskeletal & 3 & 12 \\
\hline
Renal & 3 & 12 \\
\hline
Gastrointestinal & 3 & 12 \\
\hline
Allergy & 1 & 4 \\
\hline
Geriatrics & 1 & 4 \\
\hline
Pain & 1 & 4 \\
\hline
\textbf{Total} & \textbf{30} & \textbf{120} \\
\hline
\end{tabular}
\end{table}

\section{Memory schema}

\begin{table}[h]
\centering
\small
\setlength{\tabcolsep}{9pt}
\renewcommand{\arraystretch}{1.4}
\begin{tabular}{|l|l|}
\hline
\textbf{Field} & \textbf{Meaning in this pilot} \\
\hline
feature\_type & Condition or medication \\
\hline
condition & Normalized feature label \\
\hline
status & Active, history\_of, or resolved \\
\hline
onset & Synthetic onset date or period \\
\hline
certainty & Author-assigned value from 0 to 1 \\
\hline
source\_form & Originating synthetic form type \\
\hline
negated & Whether the feature is explicitly absent \\
\hline
note & Optional synthetic context \\
\hline
\end{tabular}
\end{table}

Arm B serialized these fields into a clinical-memory block appended to the same vignette used in Arm A. The schema is a research artifact rather than a clinical interoperability standard.

\section{Reproducibility record}

\begin{itemize}
\item Benchmark version: v1.1, generated October 12, 2026.
\item Cases: 30; questions: 120; paired model calls: 60.
\item Question mix: 56 choice, 36 binary probability, 28 ordinal score.
\item Model service: Jev early-access application programming interface; model revision not recorded in the result file.
\item Primary run date: October 12, 2026.
\item Data status: entirely synthetic; no protected health information.
\end{itemize}

\section*{References}
\begin{enumerate}
\item D. Almeida. ``Introducing System One Models \& Jev.'' TypeSafe AI Blog, 2026. \url{https://typesafe.ai/blog/introducing-system-one-models-and-jev}
\item C. Packer, S. Wooders, K. Lin, V. Fang, S. G. Patil, I. Stoica, and J. E. Gonzalez. ``MemGPT: Towards LLMs as Operating Systems.'' \href{https://arxiv.org/abs/2310.08560}{\nolinkurl{arXiv:2310.08560}}, 2023.
\item G. K. Savova, J. J. Masanz, P. V. Ogren, J. Zheng, S. Sohn, K. C. Kipper-Schuler, and C. G. Chute. ``Mayo clinical Text Analysis and Knowledge Extraction System (cTAKES): architecture, component evaluation and applications.'' \textit{Journal of the American Medical Informatics Association}, 17(5):507-513, 2010. \href{https://doi.org/10.1136/jamia.2009.001560}{\nolinkurl{doi:10.1136/jamia.2009.001560}}.
\item G. W. Brier. ``Verification of Forecasts Expressed in Terms of Probability.'' \textit{Monthly Weather Review}, 78(1):1--3, 1950. \href{https://doi.org/10.1175/1520-0493(1950)078<0001:VOFEIT>2.0.CO;2}{\nolinkurl{doi:10.1175/1520-0493(1950)078<0001:VOFEIT>2.0.CO;2}}.
\end{enumerate}

\end{document}